\documentclass[preprint,12pt]{elsarticle}

\usepackage{url}
\usepackage{graphicx}
\usepackage{natbib}          % Bibliography
\usepackage{lineno}          % Line numbering
\usepackage[utf8]{inputenc}  % UTF-8 encoding
\usepackage{color}           % Colored text
\usepackage{booktabs}        % Nice-looking tables
\usepackage{multicol}        % Multiple columns  
\usepackage{amsmath}         % Math typesetting
\usepackage{amsfonts}        % Blackboard math symbols
\usepackage{amssymb}         % More blackboard math symbols
\usepackage{multirow}        % Multi-rows for tables

\newcommand{\mc}{\mathcal}
\newcommand{\M}{\mathcal{M}}

\newcommand{\R}{\mathbb{R}}
\newcommand{\STAB}[1]{\begin{tabular}{@{}c@{}}#1\end{tabular}}

\DeclareMathOperator{\arccosh}{arccosh}

\journal{Applied Soft Computing}

\begin{document}
\begin{frontmatter}
\title{Adversarial Autoencoders with Constant-Curvature Latent Manifolds}

\author[usi]{Daniele Grattarola\corref{cor1}}
\ead{daniele.grattarola@usi.ch}
\author[man,exe]{Lorenzo Livi}
\author[usi,poli]{Cesare Alippi}

\cortext[cor1]{Corresponding author}
\address[usi]{Università della Svizzera italiana, Lugano, Switzerland}
\address[man]{University of Manitoba, Winnipeg, Canada}
\address[exe]{University of Exeter, Exeter, United Kingdom}
\address[poli]{Politecnico di Milano, Milan, Italy}

\begin{abstract}
Constant-curvature Riemannian manifolds (CCMs) have been shown to be ideal embedding spaces in many application domains, as their non-Euclidean geometry can naturally account for some relevant properties of data, like hierarchy and circularity.
In this work, we introduce the \textit{CCM adversarial autoencoder} (CCM-AAE), a probabilistic generative model trained to represent a data distribution on a CCM. Our method works by matching the aggregated posterior of the CCM-AAE with a probability distribution defined on a CCM, so that the encoder implicitly learns to represent data on the CCM to fool the discriminator network. The geometric constraint is also explicitly imposed by jointly training the CCM-AAE to maximise the membership degree of the embeddings to the CCM.
While a few works in recent literature make use of either hyperspherical or hyperbolic manifolds for different learning tasks, ours is the first unified framework to seamlessly deal with CCMs of different curvatures.
We show the effectiveness of our model on three different datasets characterised by non-trivial geometry: semi-supervised classification on MNIST, link prediction on two popular citation datasets, and graph-based molecule generation using the QM9 chemical database.
Results show that our method improves upon other autoencoders based on Euclidean and non-Euclidean geometries on all tasks taken into account.
\end{abstract}

\end{frontmatter}

% \linenumbers

%%%%%%%%%%%%%%%%%%%%%%%%%%%%%%%%%%%%%%%%%%%%%%%%%%%%%%%%%%%%%%%%%%%%%%%%%%%%%%
\section{Introduction}
\noindent Many works in recent literature have highlighted that non-Euclidean geometry can naturally arise in many application domains, with \textit{constant-curvature Riemannian manifolds} (CCMs), e.g., hyperspherical and hyperbolic manifolds, playing a prominent role as embedding spaces for a variety of data distributions.
Among these, representation learning models for images \cite{davidson2018hyperspherical}, hierarchical data structures like trees and text \cite{nickel2017poincare,ganea2018hyperbolic}, relational networks \cite{gulcehre2018hyperbolic}, brain functional connectivity networks \cite{grattarola2018learning}, and dissimilarity-based datasets \cite{wilson2014spherical} have been shown to benefit from non-Euclidean embedding manifolds. 
Analysing non-Euclidean data via unsupervised deep learning, in particular, has been object of study by several recent works, with interesting results from both theoretical and practical perspectives \cite{bronstein2016geometric}. Among the most recent works studying the latent space of autoencoders we cite \cite{korman2018autoencoding,shao2017riemannian,arvanitidis2017latent}, highlighting the non-Euclidean nature of the unsupervised representations learned on various datasets.
In the more specific area of CCMs, \cite{davidson2018hyperspherical,2018arXiv180810805X} propose hyperspherical variational autoencoders to better model data on a hypersphere, whereas \cite{grattarola2018learning,zambon2018anomaly} exploit CCMs with different curvatures to perform change detection on sequences of graphs. Several other works, however, show how different types of data can greatly benefit from being embedded on CCMs, with literature going as far back as \cite{fisher1993statistical}, and the more recent notable contributions of \cite{nickel2017poincare} and \cite{wilson2014spherical}. 
However, since the application of non-Euclidean geometry to deep representation learning is a fairly recent development in machine learning \cite{bronstein2016geometric}, most methods are specific to only one type of CCM (i.e., either hyperbolic or hyperspherical) and a unified model dealing with the general family of CCMs is still missing. 

In this paper, we propose a general framework for embedding a data distribution on CCMs. The proposed solution uses adversarial learning to impose a soft constraint on the latent space of an autoencoder, training the network to produce a representation on the CCM. This work builds on the autoencoder model introduced in \cite{grattarola2018learning}, which uses adversarial learning to map a stream of graphs on a CCM for performing change detection. Here, we give a novel formulation of the model that is not restricted to operate on sequences of graphs and show its effectiveness on different learning tasks not considered in \cite{grattarola2018learning}. The model introduced here is characterised by a novel learning scheme that consists of optimising a combination of two objectives during the regularisation training phase of an adversarial autoencoder \cite{makhzani2016adversarial}.
First, we match the aggregated posterior of the autoencoder's latent representation with a prior distribution defined on the CCM. Second, in order to further impose the geometric constraint on the latent space, we train the encoder to maximise the (non-parametric) membership function of the embeddings to the CCM.
In the experimental section, we consider three tasks: semi-supervised classification on the MNIST dataset, link prediction on citation networks and molecule generation using the QM9 dataset.

The rest of the article is structured as follows: in Section \ref{sec:aae} we briefly introduce adversarial autoencoders; in Section \ref{sec:ccm_aae} we introduce our methodology and discuss some of its advantages w.r.t.\ other equivalent state-of-the-art solutions; in Section \ref{sec:exp} we provide details of the three different application scenarios that we considered in our study and discuss experimental results; in Section \ref{sec:conclusions} we summarise our contribution and provide some future research directions.

%%%%%%%%%%%%%%%%%%%%%%%%%%%%%%%%%%%%%%%%%%%%%%%%%%%%%%%%%%%%%%%%%%%%%%%%%%%%%%
\section{Background}

\subsection{Adversarial autoencoders}
\label{sec:aae}
Adversarial autoencoders (AAEs) are probabilistic models for performing variational inference, based on the framework of generative adversarial networks (GANs) \cite{goodfellow2014generative}. In AAEs the encoder network of an autoencoder is used as generator, and the aggregated posterior of the latent representation of the network is matched with an arbitrary prior distribution, by training the encoder to fool a discriminator network.

Training of AAEs occurs in two phases, namely \textit{reconstruction} and \textit{regularisation}. During the former phase, the AE is trained to reconstruct samples from the data distribution. During the regularisation phase, the discriminator is trained to distinguish between samples coming from the encoder and those coming from the true prior. Finally, the encoder is updated to fool the discriminator.
The repetition of these training steps results in a \textit{min-max} game between the encoder and the discriminator \cite{goodfellow2014generative}, where both networks compete against each other to improve at their respective tasks. 

Let $E(x)$ be the encoder network of the AAE, $D(z)$ the decoder network, $C(z)$ the critic network, $p_\textrm{data}(x)$ the data distribution on which the AAE is trained for the reconstruction phase, and finally $p(z)$ the prior distribution used for regularising the representation. 
At first, during the reconstruction phase, the AAE is trained to minimize some loss function (e.g., the mean squared error) between the input data $x \sim p_\textrm{data}(x)$ and the output of the network, $D(E(x))$. 
Then, during the regularisation phase, the adversarial optimisation can be formulated as in \cite{makhzani2016adversarial}, i.e.:
\begin{linenomath*}
\begin{equation}
    \label{eqn:aae}
    \min\limits_{E} \max\limits_{C} \mathbb{E}_{z \sim p(z)} \left[ \log C(z) \right] + \mathbb{E}_{x \sim p_\textrm{data}(x)} \left[ \log (1- C(E(x)) \right]
\end{equation}
\end{linenomath*}
The two training steps are then repeated iteratively until convergence. 

AAEs are intuitively similar to variational autoencoders (VAEs), with the key difference that AAEs replace the Kullback-Leibler divergence penalty of VAEs with the adversarial training procedure outlined above. However, his means that AAEs do not need an exact functional form of the prior in order to perform backpropagation, but only a way to sample from the prior. This makes them more flexible in the choice of prior, as it was also originally discussed by Makhzani \textit{et al.} in \cite{makhzani2016adversarial}.
In this work, we leverage this property of AAEs to constrain the latent space of the network to a CCM through a custom prior. 

\subsection{Constant-Curvature Manifolds} 
A $d$-dimensional CCM $\M$ is a Riemannian manifold characterised by a constant sectional curvature $\kappa\in\R$. 
We consider an extrinsic representation of $\mc{M}$ in its ambient space and define the CCM as 
\begin{linenomath*}
\begin{equation}
    \label{eqn:ccm}
    \mc{M} = \{x \in \R^{d+1} | \langle x, x \rangle = \kappa^{-1}\}.
\end{equation}
\end{linenomath*}

The scalar product $\langle \cdot, \cdot \rangle$ in \eqref{eqn:ccm} defines the geometry of the CCM. For $\kappa > 0$, the geometry is said to be spherical, and it is defined by the inner product: 
\begin{linenomath*}
\begin{equation}
    \label{eqn:sph_dist}
    \langle x, y \rangle = xy^T.
\end{equation}
\end{linenomath*}
In this case, the geodesic distance between points is computed using \eqref{eqn:sph_dist} as $\rho(x, y) = \arccos(\langle x, y \rangle)$.
For $\kappa < 0$, the geometry is said to be hyperbolic, and the formulation of Equation \eqref{eqn:ccm} provides the \textit{hyperboloid} model. The geometry in this case is defined from the pseudo-Euclidean scalar product:
\begin{linenomath*}
\begin{equation}
    \label{eqn:hyp_dist}
    \langle x, y\rangle = x^T \,
    \begin{pmatrix}
    I_{d\times d} & 0 \\ 0 & -1
    \end{pmatrix} \,y.
\end{equation}
\end{linenomath*}
Geodesics are computed from \eqref{eqn:hyp_dist} as $\rho(x, y) = \arccosh(\langle x, y \rangle)$.

\subsection{Priors on CCMs} 
\label{sec:ccm_prior}
The methodology presented in this paper is based on the concept of probability distributions on CCMs, which are essential to train the proposed autoencoder and impose a geometric constraint on the representation.

Let $P_{\M}(\theta)$ be a probability distribution with support on $\M$ and parametrised by vector $\theta$. Given the tangent plane $T_{x}\M \in \R^d$ at $x$, a general approach to compute $P_{\M}(\theta)$ is to take a probability distribution $P(\theta)$ with support on $T_{x}\M$, and compute $P_{\M}(\theta)$ as the push-forward distribution of $P(\theta)$ through the Riemannian exponential map (exp-map) $\mathrm{Exp}_x(\cdot)$ \cite{straub2015dirichlet,wilson2014spherical}.
Intuitively, a sample from $P_\M(\theta)$ is obtained by first sampling from $P(\theta)$ and then mapping the sample to $\M$ using the exp-map.
This provides a way to compute a distribution on a CCM starting from any distribution on the Euclidean tangent space, but in general one may use any known prior with support on $\mc{M}$. For instance, mapping the uniform distribution to the spherical manifold via exp-map may lead to counter-intuitive results, whereas one correct way of computing a spherical uniform distribution it is to orthogonally project samples from a Gaussian distribution in the ambient space onto the hypersphere.

%%%%%%%%%%%%%%%%%%%%%%%%%%%%%%%%%%%%%%%%%%%%%%%%%%%%%%%%%%%%%%%%%%%%%%%%%%%%%%
\section{Adversarial Autoencoders on CCMs}
\label{sec:ccm_aae}
\begin{figure*}
    \centering
    \includegraphics[width=\textwidth, keepaspectratio=true]{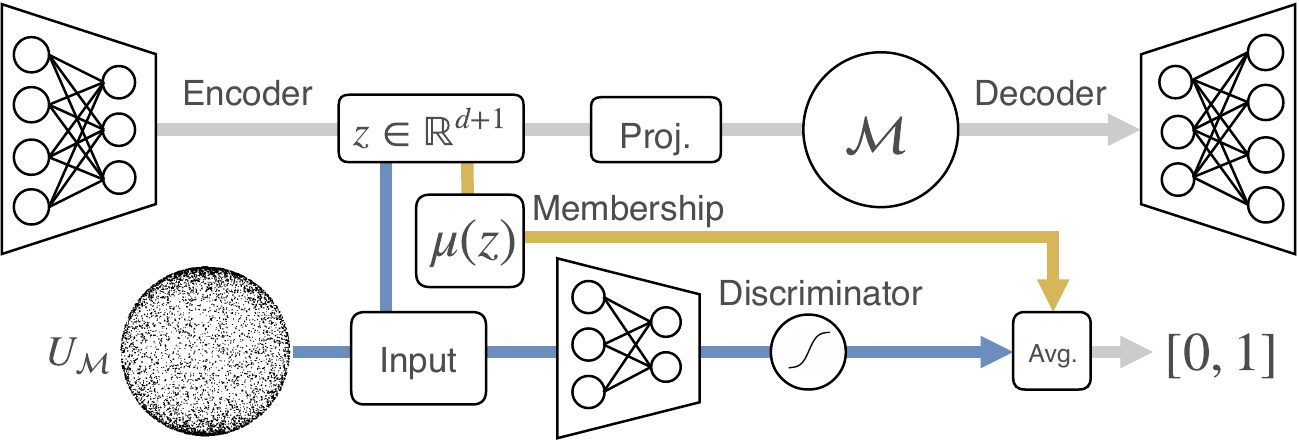}
    \caption{Schematic view of the spherical CCM-AAE. From left to right: the encoder produces embeddings $z \in \R^{d+1}$ in the ambient space, which are (optionally) projected onto the CCM before being fed to the decoder. The discriminator is trained to distinguish between embeddings and samples coming from the spherical uniform prior (blue path). Finally, the membership degree of the embeddings (yellow path) is averaged with the discriminator in order to compute the loss and update the encoder. Best viewed in colour.}
    \label{fig:scheme}
\end{figure*}
We consider the problem of mapping a data distribution to a $d$-dimensional CCM $\M$ with curvature $\kappa \in \R$, as well as learning a map from $\M$ to the data space in order to generate new samples. 
Using adversarial learning, the latent representation of the CCM-AAE is matched to a prior distribution defined on the CCM, while jointly training the encoder to produce embeddings that lie on the CCM via an explicit regularisation term in the loss, penalising the encoder when the embeddings are far from the manifold.
This facilitates the network in converging to the target manifold, making it easier to match the aggregated posterior to the prior. A schematic view of the proposed architecture is shown in Figure \ref{fig:scheme}.

The methodology presented here is independent from the type of neural network used to learn the representation of the data, and can in principle be applied as a general regularisation technique for different tasks. In this section, we provide a general outline of the approach, and leave the task-specific details of implementation to the experimental section. 

\subsection{Method} 
Using the same notation adopted in Section \ref{sec:aae}, we denote with $E(x)$ the encoder network, $D(z)$ the decoder, $C(z)$ the discriminator, and $p_\textrm{data}(x)$ the data distribution, where samples $x \sim p_\textrm{data}(x)$ are in some input space $\mathcal{X}$. 
The CCM-AAE is then defined as the composition of two maps: 
\begin{itemize}
    \item $E: \mc X \rightarrow \mc M$, mapping data to the manifold;
    \item $D: \mc M \rightarrow \mc X$, mapping embeddings back the data space.
\end{itemize}

In practice, the latent space of the autoencoder is taken as $\R^{(d+1)}$ and represents the ambient space of $\mc M$.

During the reconstruction phase, the autoencoder is trained as usual to reconstruct samples from the data distribution, minimising a loss function between $x \in \mc X$ and $D(E(x))$. 
During the regularisation phase, we train the critic network to discriminate between samples coming from the encoder and samples from the true prior $P_\M(\theta)$, and then we update the encoder to fool the critic network. 
By matching the posterior to $P_\M(\theta)$, the network is implicitly constrained to embed input data on the CCM, and the solution to the adversarial game can be obtained from Equation \eqref{eqn:aae} as:
\begin{linenomath*}
\begin{equation}
    \label{eqn:ccm_aae_no_memb}
    \min\limits_{E} \max\limits_{C} \mathbb{E}_{z \sim P_\M(\theta)} \left[ \log C(z) \right] + \mathbb{E}_{x \sim p_\textrm{data}(x)} \left[ \log (1- C(E(x)) \right]
\end{equation}
\end{linenomath*}

However, this implicit optimisation is often not sufficient for the network to effectively learn the non-Euclidean geometry of the CCM, as also highlighted by the experimental results shown in \cite{grattarola2018learning}. Consequently, here we also train the encoder network to maximise the membership degree of the embeddings to $\M$, so that the loss landscape is explicitly modified in favour of those embeddings that lie exactly on the CCM. 
\begin{figure}
    \centering
    \includegraphics[width=0.8\linewidth, keepaspectratio=true]{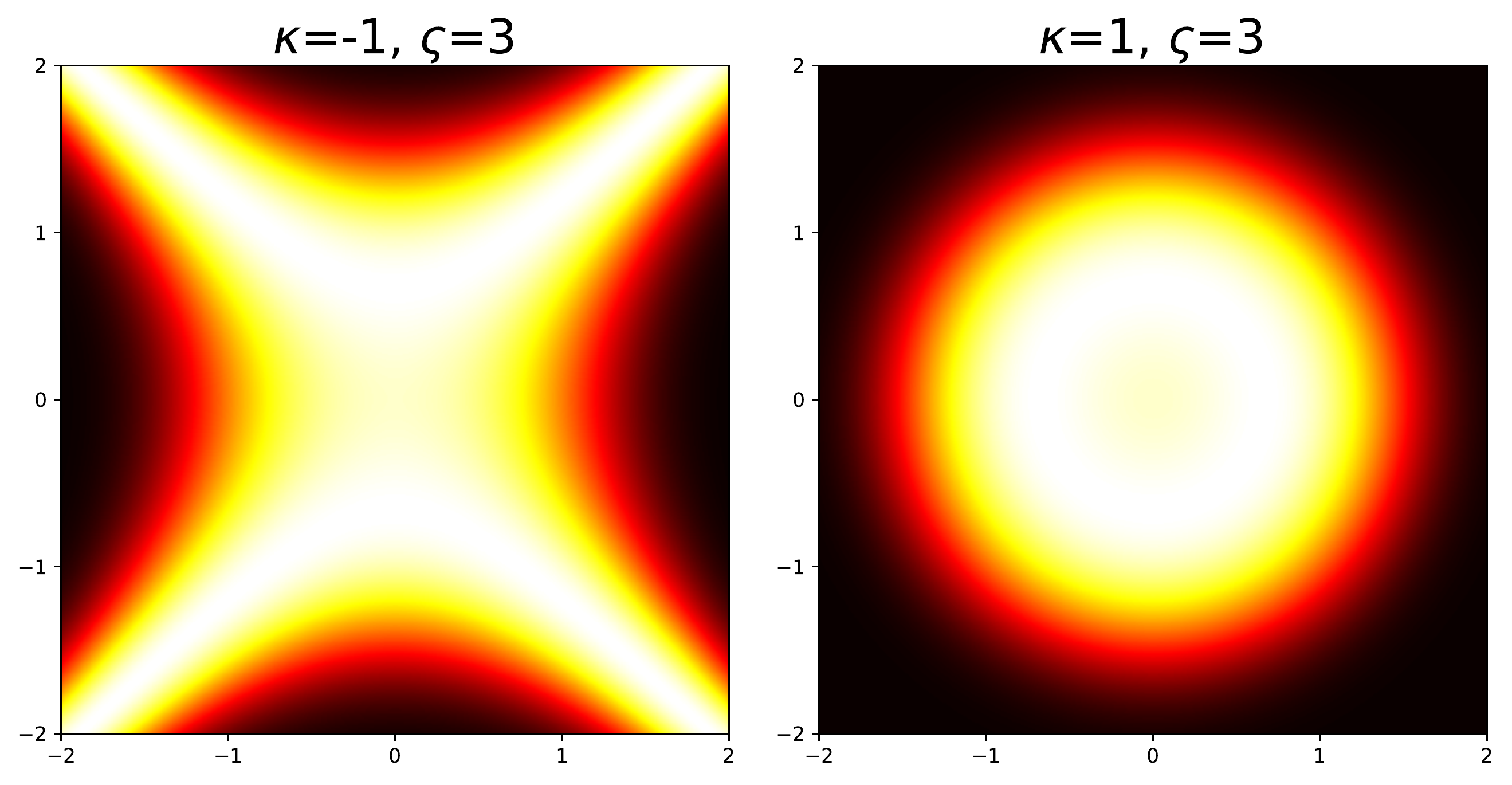}
    \caption{Membership function of a hyperbolic (left) and a spherical (right) CCM in the case of $d=1$. Lighter colours indicate higher values (white = 1; best viewed in colour).}
    \label{fig:membership}
\end{figure}
For a manifold $\M$ with $\kappa \ne 0$, the membership degree of a sample $z$ is computed as (see Figure \ref{fig:membership}): 
\begin{linenomath*}
\begin{equation}
    \label{eqn:membership}
    \mu(z) = \mathrm{exp}\left(\frac{-\big( \langle z, z \rangle - \frac{1}{\kappa} \big)^2}{2\varsigma^2}\right)
\end{equation}
\end{linenomath*}
where $\varsigma \ne 0$ controls the width of the membership function. 
In practice, we optimise both regularisation objectives in parallel, by taking the average of the critic's output and the membership degree of the embeddings when updating the encoder in the regularisation phase as: 
\begin{linenomath*}
\begin{equation}
    \tilde{C}(z) = \frac{C(z) + \mu(z)}{2}.
\end{equation}
\end{linenomath*}

The final form of the regularisation for the CCM-AAE can then be written as: 
\begin{linenomath*}
\begin{equation}
    \label{eqn:ccm_aae}
    \min\limits_{E} \max\limits_{C} \mathbb{E}_{z \sim P_\M(\theta)} \left[ \log \tilde{C}(z) \right] + \mathbb{E}_{x \sim p_\textrm{data}(x)} \left[ \log (1- \tilde{C}(E(x)) \right].
\end{equation}
\end{linenomath*}

\paragraph{Projection to the CCM} When exact operations need to be computed on the manifold (e.g., distances, sampling), we compute an orthogonal projection of the embeddings onto $\M$ in order to account for the inevitable error in the model. 
The projection can be embedded as a layer in the network or computed only at test time. For instance, for KNN-based semi-supervised classification, where we need to compute the pairwise geodesic distances of the embeddings, we first let the network learn a representation and then project the embeddings onto $\M$ to compute similarities at test time.
Alternatively, when using the CCM-AAE for generating new samples in the data space, we ensure that the decoder network learns a meaningful map between $\M$ and the data space by always projecting the latent codes onto $\M$. This does not impact the regularisation of the encoder, and has only a marginal effect on the convergence of the network.

\subsection{Advantages} 
A key difference of our approach with other works in the literature is that we do not impose the non-Euclidean geometry on the latent space by simply projecting the embeddings onto the CCM, or otherwise explicitly limiting the latent space (e.g., by sampling embeddings from the CCM prior \cite{davidson2018hyperspherical}). The encoder has to learn the latent manifold autonomously, because the projection is not performed during the regularisation step.
Moreover, as highlighted in previous sections, AAEs have the advantage w.r.t.\ to VAEs of not requiring the explicit form of the prior in order to perform backpropagation. This is especially relevant when dealing with non-Euclidean geometry, where functional forms can be analytically complex, as any valid prior on an Euclidean tangent space can be a suitable prior for the network (via exp-map).
A relevant effect of this is discussed in the experimental section, where we show that the spherical version of our model is able to deal with high-dimensional manifolds better than an equivalent VAE with spherical latent space. This results in a more stable performance when using high-dimensional manifolds on the considered applications (we show a specific example of this on a semi-supervised classification task on MNIST), as our model does not suffer from the same performance drop denoted by the spherical VAE.

%%%%%%%%%%%%%%%%%%%%%%%%%%%%%%%%%%%%%%%%%%%%%%%%%%%%%%%%%%%%%%%%%%%%%%%%%%%%%%
\section{Experiments}
\label{sec:exp}
We perform experiments to validate our methodology in three relevant settings. First, following the experimental setting of Davidson \textit{et al.\ }\cite{davidson2018hyperspherical}, we report a performance comparison of different models on semi-supervised image classification on MNIST and link prediction on citation networks. While the link prediction task requires to compute embeddings of the citation networks only at the node level (i.e., the representation is learned for each node of a single graph), for the third experiment we also evaluate our methodology on a task that requires to process graphs as individual inputs. In this case, we consider the problem of graph-based molecule generation, because it represents an interesting open challenge that has attracted the interest of the machine learning community \cite{simonovsky2018graphvae,de2018molgan}. 
All of the considered settings have been shown to benefit from non-Euclidean representations of the data \cite{davidson2018hyperspherical,grattarola2018learning}, and therefore provide a good platform for testing our method and comparing it to previous literature. 

For each experiment, we test the two main configurations of CCM-AAE, i.e., with spherical and hyperbolic geometry. The geometry of the CCM is dependent only on the sign of the curvature, whereas the absolute value of the curvature has only an effect on the scale of the representation. For this reason, and in order to simplify the implementation of the CCM-AAE, here we only consider $\kappa = 1$ and $\kappa = -1$.

For the prior, we follow Makhzani \textit{et al.\ }\cite{makhzani2016adversarial}, and adapt the standard normal distribution $\mc N(0,1)$ to our setting.
For $\kappa = -1$ we use the push-forward standard normal $\mc{N}_\mc{M}(0, 1)$, where the origin of the exp-map is taken as the point $x \in R^{d+1}$ such that $x_i = 0, i=1, \dots, d$ and $x_{d+1} = 1$ (the point is chosen to simplify some implementation details, but any other point on the manifold would be suitable).

For $\kappa = 1$ however, as the dimension $d$ of the manifold grows, mapping $N(0, 1)$ to the CCM via exp-map quickly results in a uniform distribution on the sphere. A similar consideration was also noted in \cite{davidson2018hyperspherical} for the von Mises-Fisher distribution. Therefore, a better choice is to use directly the spherical uniform distribution as prior (c.f.\ Section \ref{sec:ccm_prior}).

All experiments were conducted on a machine with an Intel Core i7 CPU with 4 cores and 8 threads, 16GB of RAM, and an Nvidia Titan Xp GPU with 12GB of VRAM, using the Keras library as high-level interface to TensorFlow. The average duration of training epochs across the three experiments is reported in Table \ref{tab:epochs}. Note that training times for $\kappa = -1$ are substantially higher than in the spherical case, because sampling from a hyperbolic space requires computing the exp-map of the samples for each batch. On the other hand, sampling from a uniform distribution on a spherical manifold is much faster due the implementation described in Section \ref{sec:ccm_prior}.
\begin{table}[]
    \centering
    \caption{Average duration of training epochs of the CCM-AAE on the hardware used for experiments. We report the duration for the two tested values of $\kappa$ across all three experiments. For link prediction, the value refers to the duration on the Cora dataset. For molecule generation, the value refers to the CCM-AAE without graph matching.}
    \vspace{1em}
    \begin{tabular}{l|c|c}
        \toprule
        \textbf{Experiment} & $\kappa$ & s / epoch \\
        \midrule
        \multirow{2}{*}{MNIST} & $1$  & 1.73 \\
                               & $-1$ & 5.29 \\
        \midrule
        \multirow{2}{*}{Link pred.} & $1$  & 0.11 \\
                                    & $-1$ & 0.32 \\
        \midrule
        \multirow{2}{*}{Molecule gen.} & $1$  & 35.69 \\
                                       & $-1$ & 60.05 \\
        \bottomrule
    \end{tabular}
    \label{tab:epochs}
\end{table}

%%%%%%%%%%%%%%%%%%%%%%%%%%%%%%%%%%%%%%%%%%%%%%%%%%%%%%%%%%%%%%%%%%%%%%%%%%%%%%
\subsection{Semi-supervised Image Classification}
\begin{table}
\centering
\caption{Accuracy of semi-supervised K-NN classification on MNIST for 100, 600, and 1000 observed training labels per class w.r.t. the dimensionality of the latent manifold. We report mean and standard deviation computed over 10 runs.}
\vspace{1em}
\begin{tabular}{@{}l|c|ccc@{}}
\toprule
\textbf{Method} & \textbf{d} & \textbf{l=100} & \textbf{l=600} & \textbf{l=1000} \\ \midrule
VAE & 10 & 89.1 {\tiny $\pm 0.6$} & 92.7 {\tiny $\pm 0.5$} & 93.3 {\tiny $\pm 0.5$} \\
$\mathcal{S}$-VAE & 10 & 90.7 {\tiny $\pm 0.7$} & 93.7 {\tiny $\pm 0.5$} & 94.1 {\tiny $\pm 0.5$} \\
AAE & 100 & 91.2 {\tiny $\pm 0.5$} & 94.9 {\tiny $\pm 0.2$} & 95.4 {\tiny $\pm 0.2$} \\
Ours ($\kappa = 1$) & 20 & 91.4 {\tiny $\pm 0.4$} & 95.0 {\tiny $\pm 0.5$} & 95.6 {\tiny $\pm 0.3$} \\
\textbf{Ours ($\kappa = -1$)} & 30 & \textbf{91.5 {\tiny $\pm 0.3$}} & \textbf{95.2 {\tiny $\pm 0.2$}} & \textbf{95.8 {\tiny $\pm 0.2$}} \\ \bottomrule
\end{tabular}
\label{tab:mnist_results}
\end{table}

Following the methodology of \cite{kingma2014semi}, we evaluate the quality of the embeddings produced by the CCM-AAE on a semi-supervised classification task on dynamically binarised MNIST \cite{salakhutdinov2008quantitative}. We train the CCM-AAE on a random split of 55k samples for training, 10k for testing, and 5k for validation and model selection. \\
After training, we draw for each class $l = 100, 600, 1000$ pairs of samples and labels uniformly from the training set, and evaluate the test accuracy of a K-NN classifier on the embeddings produced by hyperbolic and hyperspherical CCM-AAEs.

\paragraph{Setting} Similarly to the experimental setting of \cite{davidson2018hyperspherical}, the encoder is a two-layer, fully connected network of 256 and 128 neurons with ReLU activations, followed by a linear layer with $d+1$ neurons to produce the latent representation. The decoder has two ReLU layers with 128 and 256 units, followed by an output layer with sigmoid activations.

\begin{table}[]
    \centering
    \caption{Hyperparameter configuration of the CCM-AAE for MNIST. The \textit{Searched} columns indicates that the final value was found via grid search among the indicated values, using the validation loss for model selection. Alternatively, when we did not perform a grid search, we indicate how the value was found (\textit{Keras default} indicates that the value was the default setting for the popular deep learning library Keras, which we used for experiments).
    An emtpy final value in the \textit{Value} column indicates that the grid search was repeated for each combination of dataset and $\kappa$.}
    \vspace{1em}
    \begin{tabular}{l|c|l}
        \toprule
        \textbf{Hparam.} & \textbf{Value} & \textbf{Searched} \\
        \midrule
        $d$ & - &  2, 5, 10, 20, 40, 60, 100 \\
        \midrule
        $h$ & 64 &  32, 64, 128 \\
        LeakyReLU $\alpha$ & 0.3 & Keras default \\
        L2 reg. (for $\tilde{C}(z)$) & 0.01 & Keras default \\
        \midrule
        $\varsigma$ (for $\mu(z)$) & 5 & 1, 2, 5, 10 \\
        \midrule
        Learning rate & 0.001 & $0.001$, $0.005$, $0.01$ \\
        Batch size & 1024 & Empirically \\
        \bottomrule
    \end{tabular}
    \label{tab:hparams_mnist}
\end{table}

For hyperparameters specific to the CCM-AAE (and other hyperparameters), we perform a brief grid search using the validation loss to perform model selection. The tested values and final configuration are reported in Table \ref{tab:hparams_mnist}. 
We adopt a fully connected discriminator with two layers of $h$ units, leaky ReLUs, and L2 regularisation, followed by an output layer with sigmoid activation. 
We train both networks using Adam until convergence, using early stopping on the autoencoder's validation loss with a look-ahead of 50 epochs (value taken from \cite{davidson2018hyperspherical}). For both networks we optimise the cross-entropy loss between the inputs and reconstructed images. 

The embeddings of the network are exactly projected onto the manifold only at test time, to compute the mutual geodesic distances for K-NN (for which we set $K = 5$ as in \cite{davidson2018hyperspherical}).
We compare our results using the same network architecture and configuration to train a standard adversarial autoencoder with Gaussian prior (AAE) \cite{makhzani2016adversarial}, a variational autoencoder with Gaussian prior (VAE) \cite{kingma2013auto}, and the hyperspherical variational autoencoder proposed by \cite{davidson2018hyperspherical} ($\mc{S}$-VAE). For $\mc{S}$-VAE, we use the open source implementation provided by the authors in the original paper\footnote{https://github.com/nicola-decao/s-vae-tf}.
\begin{figure}
    \centering
    \includegraphics[width=0.8\linewidth, keepaspectratio=true]{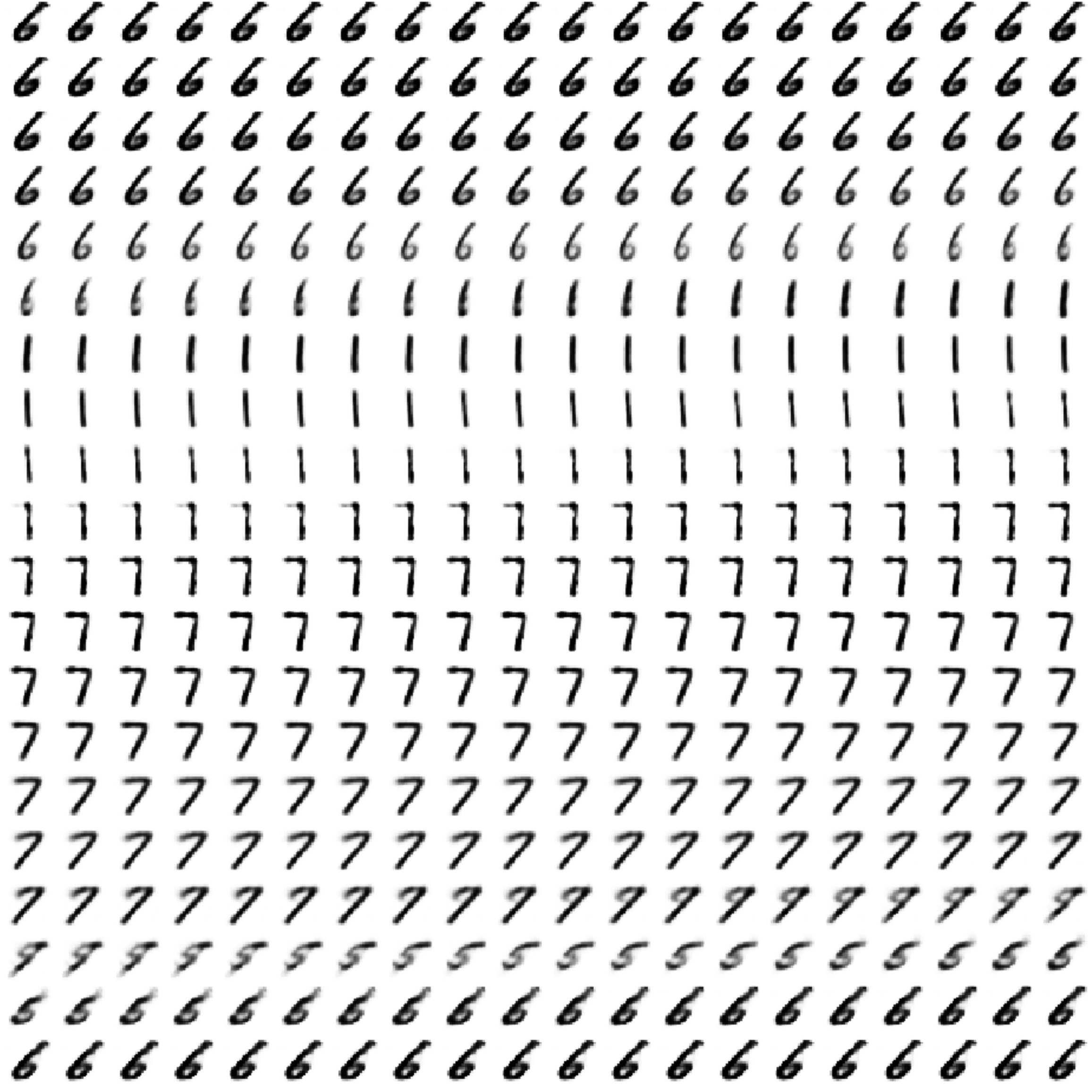}
    \caption{Traversing the latent space of a spherical CCM-AAE ($d=2$) along an equator of the sphere (samples are arranged left-to-right, top-to-bottom). Note how the digits are smoothly represented in a circular way, suggesting how the data can be naturally encoded on a sphere.}
    \label{fig:ls_mnist}
\end{figure}
\begin{figure}
    \centering
    \includegraphics[width=\linewidth, keepaspectratio=true]{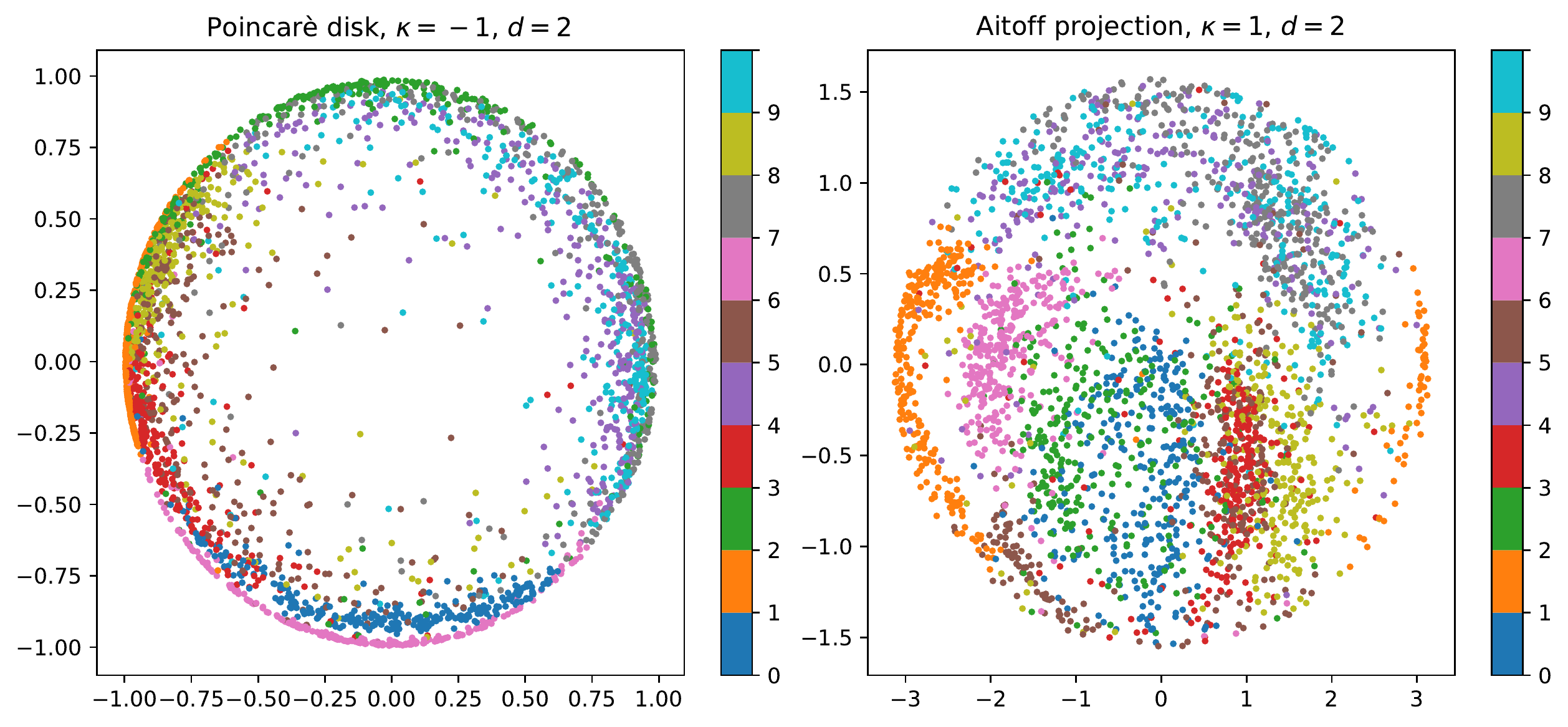}
    \caption{Embeddings produced by the CCM-AAE on MNIST, for $d=2$. We report the Poincaré disk model of the latent CCM for $\kappa = -1$, and the Aitoff projection for $\kappa = 1$.}
    \label{fig:embeddings}
\end{figure}

\paragraph{Results} The best results obtained by each method are summarised in Table \ref{tab:mnist_results}. We note that the adversarial setting consistently outperforms its variational counterparts, even when considering the non-Euclidean $\mc{S}$-VAE method.
The CCM-AAE also performs slightly better on average w.r.t. to the Euclidean AAE, with no significant statistical differences observed between spherical and hyperbolic embeddings. However, we note that the Euclidean model requires a significantly higher $d$ in order to match the performance of the non-Euclidean ones, with the spherical model being the most efficient in this regard.
This confirms an already observed fact in the literature \cite{davidson2018hyperspherical}, and a possible explanation for this phenomenon can be intuitively seen in Figure \ref{fig:ls_mnist}, where MNIST is shown to have a natural representation on a spherical domain.

Finally, we show in Figure \ref{fig:embeddings} that the CCM-AAE is able to learn a sufficiently good representation of the data even at very low dimensions (pictured for $d=2$ in order to visualise it on paper), with no substantial differences between hyperbolic and spherical geometries, as also highlighted by the performance of KNN.

%%%%%%%%%%%%%%%%%%%%%%%%%%%%%%%%%%%%%%%%%%%%%%%%%%%%%%%%%%%%%%%%%%%%%%%%%%%%%%
\subsection{Link Prediction}
\begin{table*}
\centering
\caption{Best average AUC and AP of semi-supervised link-prediction on the Cora and Citeseer datasets. We report mean and standard deviation over 5 runs. Best results are not highlighted due to the differences between the best algorithms not being statistically significant.}
\vspace{1em}
\begin{tabular}{@{}l|ccc|ccc@{}}
\toprule
 &  & \multicolumn{2}{c}{Cora} &  & \multicolumn{2}{c}{Citeseer} \\
\textbf{Method} & \textbf{d} & \textbf{AUC} & \textbf{AP} & \textbf{d} & \textbf{AUC} & \textbf{AP} \\ \midrule
VGAE & 20 & 91.8 {\tiny $\pm 0.8$} & 92.9 {\tiny $\pm 0.6$} & 16 & 90.6 {\tiny $\pm 1.3$} & 91.7 {\tiny $\pm 1.1$} \\
AAE & 64 & 93.4 {\tiny $\pm 0.6$} & 93.8 {\tiny $\pm 0.7$} & 64 & 94.0 {\tiny $\pm 0.8$} & 94.6 {\tiny $\pm 1.0$} \\
Ours ($\kappa = 1$) & 8 & 93.4 {\tiny $\pm 0.7$} & 93.9 {\tiny $\pm 0.8$} & 8 & 92.8 {\tiny $\pm 0.4$} & 93.4 {\tiny $\pm 0.4$} \\
Ours ($\kappa = -1$) & 64 & 89.4 {\tiny $\pm 0.9$} & 90.4 {\tiny $\pm 1.0$} & 8 & 91.0 {\tiny $\pm 0.4$} & 91.6 {\tiny $\pm 0.4$} \\ \bottomrule
\end{tabular}
\label{tab:lp_results}
\end{table*}

For the link prediction task, we follow the methodology of \cite{kipf2016variational} and evaluate the performance of the CCM-AAE on the popular Cora and Citeseer citation network datasets\footnote{The Pubmed dataset is often considered alongside the other two, but the high number of nodes in the network caused GPU memory issues with all tested algorithms, and we therefore do not report results for it.}.
In this task, we train the CCM-AAE to predict connections on a subset of the network, and evaluate the area under the ROC (AUC) and average precision (AP) of the model in predicting a test set of held-out links. We split the data randomly, using 10\% of the links for testing and 5\% for validation and model selection. 

\paragraph{Graph data} Cora and Citeseer are two popular network datasets representing citation links between documents, with sparse node attributes representing text features found in the documents. Each node is also associated with a class label, which we do not use here. We represent a network with $N$ nodes and $F$-dimensional node attributes as a tuple $(A, X)$, where $A \in \{0, 1\}^{N \times N}$ is the symmetric adjacency matrix of the network, and $X \in \{0, 1\}^{N \times F}$ is the node attribute matrix.
For the Cora dataset we have $N=2708$ and $F=1433$, whereas for Citeseer we have $N=3327$ and $F=3703$. The networks have an average degree of $4$ and $2.84$, respectively.

\paragraph{Setting} The CCM-AAE has the same structure of the variational graph autoencoder (VGAE) in \cite{kipf2016variational}, with a graph convolutional encoder network \cite{kipf2017semi} followed by a scalar product decoder. The encoder consists of a graph convolutional layer with 32 channels and ReLU activations, followed by a $d+1$ dimensional graph convolutional linear layer. Dropout is applied before every layer.

For the decoder, we first project the latent representation onto the target manifold, and then we reconstruct the adjacency matrix by computing the scalar product between node embeddings, followed by an activation to normalise the output between 0 and 1. 
In the spherical case, the scalar product can be interpreted as computing a cosine similarity between embeddings, which we then normalise with a sigmoid activation. For the hyperbolic CCM-AAE, the pseudo-Euclidean scalar product assumes values in the $(-\infty, -1]$ range, so we normalise it to $(0, 1]$ by applying a shifted exponential as activation to the decoder's output, $\sigma(x) = \exp(x + 1)$.
In principle, any normalisation function can be used here, but we leave further exploration of this matter as future work. For instance, an obvious way of normalising the output would be to add a final layer with sigmoid activations and let the network learn how to map the scalar product to a prediction in $(0, 1)$. However, here we wanted to have the same number of parameters across all models to report a fair comparison.\\

We keep most of the configuration used for MNIST unvaried, but we perform a grid search over the dropout rate ($0.0$, $0.1$, $0.2$, $0.3$, $0.4$) for each dataset and geometry. 
Additionally, we repeat the grid search over the dimension $d$ using similar values to those reported in \cite{davidson2018hyperspherical} ($8$, $16$, $20$, $32$, $64$, $128$).
Both networks are trained using Adam to optimise a cross-entropy loss (when training the autoencoder, we apply the same re-weighting technique used in \cite{kipf2016variational}). We train the model using early stopping on the validation AUC with a patience of 100 epochs (value taken from \cite{kipf2016variational}). 

\paragraph{Results} We report results in Table \ref{tab:lp_results}.
We compare our results against VGAE and an AAE, using the same network architecture for all models. 
The spherical CCM-AAE is able to consistently outperform VGAE in both tasks, but we highlight a drop in performance in the hyperbolic model. Once again, the Euclidean AAE performs comparably to the spherical one, but requires a significantly higher-dimensional latent space. 

\begin{figure}
    \centering
    \includegraphics[width=0.8\linewidth, keepaspectratio=true]{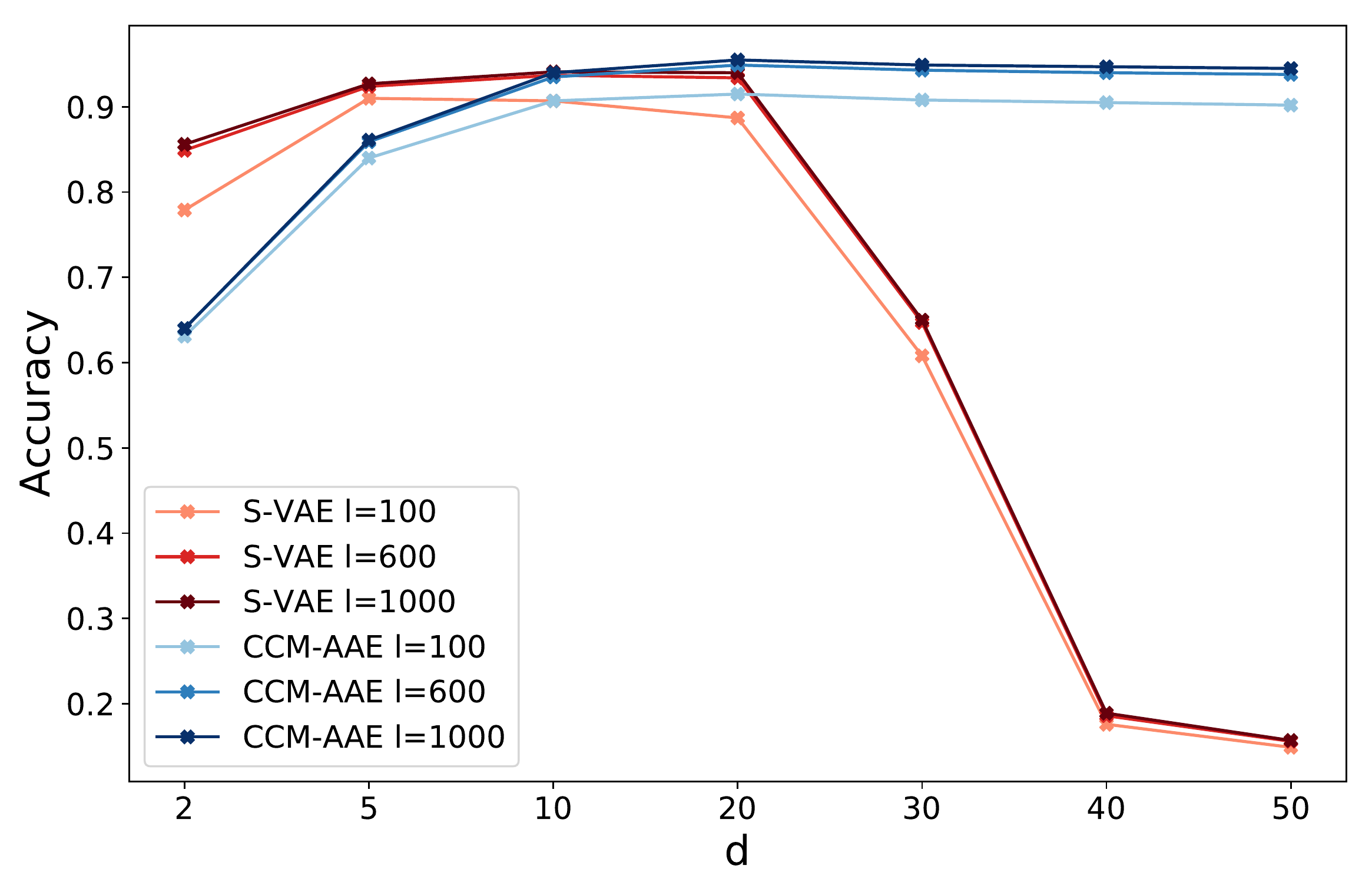}
    \caption{Comparison between the spherical CCM-AAE and $\mc{S}$-VAE considering the mean test accuracy on MNIST w.r.t. manifold dimension. $\mc{S}$-VAE denotes a performance collapse as reported by \cite{davidson2018hyperspherical}, but the CCM-AAE maintains a stable performance even at higher dimensions.}
    \label{fig:cod}
\end{figure}

While the code used to implement $\mc{S}$-VAE worked as intended on MNIST, on the link prediction task we encountered numerical issues that made it impossible to replicate the results of \cite{davidson2018hyperspherical} in our different experimental setting with different data splits, re-weighting technique for the loss, and hyperparameters searched. When further investigating the instability of $\mc{S}$-VAE, we observed a computational issue in the model, which would saturate the floating point representation of the GPU, resulting in invalid gradients being propagated through the network. The numerical instability derives from the sampling procedure of the von Mises-Fisher distribution on which the model is based, as the exponentially scaled modified Bessel function used for sampling causes a division by zero for higher values of $d$. 
On the other hand, we note that the performance of the proposed spherical CCM-AAE does not suffer from the same issue when using high-dimensional latent spaces (shown in Figure \ref{fig:cod} for MNIST).
% A possible explanation for this different behaviour is that the CCM-AAE is free to optimise the representation on the full ambient space, potentially converging to a manifold with lower curvature in order to maximise the membership $\mu(z)$ (Eq. \eqref{eqn:membership}) while keeping the representation uncollapsed.
%

%%%%%%%%%%%%%%%%%%%%%%%%%%%%%%%%%%%%%%%%%%%%%%%%%%%%%%%%%%%%%%%%%%%%%%%%%%%%%%
\subsection{Molecule Generation}
Graph-based molecule generation is a fairly recent research area, which is starting to get attention from the machine learning and cheminformatics communities \cite{simonovsky2018graphvae,de2018molgan}. 
Differently from past approaches in molecule generation, most of which relied on the SMILES string representation of molecules, the graph-based approach represents atoms as nodes in a graph, with chemical bonds represented as attributed edges.
Therefore, this novel approach considers graphs as objects in the data space (in contrast to the usual approach in graph-based deep learning where a single graph is embedded at the node level) and is a relevant application scenario for our methodology.

We compare the proposed CCM-AAE against several molecule generation models on the QM9 dataset of small molecules, and evaluate the performance of our model following the methodology in \cite{simonovsky2018graphvae}.
QM9 contains $\sim\!\!\!134$k small molecules of up to 9 heavy atoms, with 4 atomic numbers and 3 bond types. We split the data according to \cite{simonovsky2018graphvae}, using 10k samples for testing, and 10k for validation and model selection.

To evaluate the quality of the model, we sample random points from the latent CCM, and map them to the molecule space using the decoder. We then compute three metrics for the generated molecules \cite{samanta2018NeVAEAD}: the \textit{validity} measure indicates the fraction of molecules that are chemically valid, the \textit{novelty} metric indicates the fraction of valid molecules that are not in the original QM9 dataset, and finally the \textit{uniqueness} metric indicates the fraction of unique molecules among the valid ones. Finally, in order to quantify the overall performance of a model, we aggregate the three metrics by multiplying them together. If we assume independence of uniqueness and novelty given validity, this can be seen as computing the probability of generating a valid, unique, and novel molecule. We use this aggregated \textit{joint} metric to compare different models between them.
The assumption of independence among the metrics was also validated empirically, by computing the true ratio of valid, unique, and novel molecules generated by our algorithm. The difference between the true ratio and the value obtained by multiplying \textit{validity}, \textit{novelty}, and \textit{uniqueness}, was not statistically significant in the conducted experiments.

\paragraph{Molecules representation} We represent molecules as attributed graphs following the approach of \cite{simonovsky2018graphvae}. The representation is similar to the one used in the link prediction setting, with the addition of attributed edges to describe chemical bonds. We use a one-hot binary representation for the attributes, where nodes are labelled with one of 4 possible atomic numbers, edges with one of 3 possible types of bonds, and we also explicitly represent null node and edge types with a dedicated class label. Therefore, graphs are represented as tuples $(A, X, E)$, where $A \in \{0, 1\}^{N \times N}$ and $X \in \{0, 1\}^{N \times 5}$ have the same meaning described above, and $E \in \{0, 1\}^{N \times N \times 4}$ is the edge attributes matrix.
Similarly to \cite{simonovsky2018graphvae}, we include support for smaller graphs (up to $N = 9$) via zero-padding. 

\paragraph{Setting} We structure the CCM-AAE according to the same architecture of \textit{GraphVAE} \cite{simonovsky2018graphvae}, of which we consider the unconditional version for simplicity. The encoder is a graph-convolutional network based on \textit{edge-conditioned} graph convolutions \cite{simonovsky2017ecc}, composed of two layers of 32 and 64 channels, with ReLUs, batch normalisation, and a filter-generating network composed of a single linear layer. The convolutions are followed by a global gated attention pooling layer \cite{li2015gated} with 128 units, and a linear layer with $d+1$ units to map the representation to the ambient space of the CCM. 
The decoder is a fully connected network with three layers of 128, 256, and 512 neurons, with ReLUs, and batch normalisation, followed by three parallel output layers to produce the reconstructed $A$, $X$, and $E$. The first output layer has sigmoid activations, whereas the latter two have node- and edge-wise softmax activations, respectively. The embeddings are projected to the CCM before being fed to the decoder.  
The configuration of the discriminator and training procedure is again unvaried. The network is trained until convergence, monitoring the validation reconstruction loss with a look-ahead of 25 epochs (i.e., the number of epochs used to train GraphVAE in \cite{simonovsky2018graphvae}).

Following \cite{simonovsky2018graphvae}, we train the CCM-AAE using graph matching in order to account for graphs with unidentified nodes. This consists of matching the input graphs to the generated outputs before computing the loss for backpropagation, so that the network learns a permutation-invariant representation.
We apply the same max-pooling matching algorithm \cite{cho2014finding} used for GraphVAE, with 75 iterations and the same affinity function described in \cite{simonovsky2018graphvae}. 
We implemented the loss function as in \cite{simonovsky2018graphvae}, with the same tricks to speed up training: (i) we impose the symmetry of the output matrices during post-processing (by removing those edges for which $A_{ij} \ne A_{ji}$), (ii) we include in the prediction the maximum spanning tree on the set of probable nodes ($A_{ii} \ge 0.5$), and (iii) we ignore hydrogen atoms and only add them as padding during chemical validation\footnote{We used the RDKit framework for chemical validation and hydrogen padding.}. 
As final trick to speed up convergence, we apply a re-weighting of the loss function to mitigate the importance of the null nodes and edges, and to improve the reconstruction of the rarer edges. The weight is computed from the dataset-wide inverse document frequency (IDF) score of each element in $A$, $X$, and $E$. For instance, the IDF score of the adjacency matrix across a dataset $\mc{D}$ is computed as:

\[
\mathrm{IDF}_{ij} = \log{\left(\frac{|\mc{D}|}{1+ \sum_{k}^{|\mc{D}|} A_{ij}^{(k)}}\right)}
\]

and the log-loss between $A$ and its reconstruction $A'$ is reweighted as: 

\[
\log(A'|A) = \sum\limits_{i, j} \left(1 + A_{ij}\mathrm{IDF}_{ij}\right) \left( A'_{ij} \log A_{ij} + (1-A'_{ij})\log (1-A_{ij}) \right)
\]

In order to account for the node permutations, the $\mathrm{IDF}$ weight matrices are matched to their respective target matrices before computing the loss.

We report our results along the others of several models for molecule generation, namely GraphVAE, \textit{MolGAN} \cite{de2018molgan}, the character-based CVAE \cite{gomez2018automatic}, and the grammar-based GVAE \cite{kusner2017grammar} (the latter two use SMILES representations). Comparisons are reported for algorithms that include the graph matching step, as well as those that do not, and we report the performance of our model in both cases. 
\begin{table}
\centering
\caption{Validity, uniqueness, and novelty metrics on QM9, with and without graph matching. The ``Joint'' column shows the aggregated score computed as the product of the three metrics, providing a general idea of the overall performance of the models. Baseline results are taken from \cite{simonovsky2018graphvae,de2018molgan}. The best individual metrics and the model with the best aggregated score are highlighted in bold.}
\vspace{1em}
\begin{tabular}{@{}ll|cccc|c@{}}
\toprule
 & \textbf{Method} & \textbf{d} & \textbf{Valid} & \textbf{Uniq.} & \textbf{Novel} & \textbf{Joint} \\ \midrule
\multirow{6}{*}{\STAB{\rotatebox[origin=c]{90}{\textbf{No match.}}}} & GraphVAE & 80 & 81.0 & 61.0 & 24.1 & 11.9 \\
 & CVAE & 60 & 10.3 & 67.5 & 90.0 & 6.3 \\
 & GVAE & 20 & 60.2 & 9.3 & 80.9 & 4.5 \\
 & MolGAN & - & \textbf{98.1} & 10.4 & \textbf{94.2} & 9.6 \\
 & AAE & 80 & 30.1 & \textbf{92.7} & 84.8 & 23.7 \\
 & \textbf{Ours ($\kappa = 1$)} & 80 & 36.3 & 92.6 & 87.1 & \textbf{29.2} \\
 & Ours ($\kappa = -1$) & 80 & 22.5 & 86.1 & 70.2 & 13.6 \\ \midrule

\multirow{5}{*}{\STAB{\rotatebox[origin=c]{90}{\textbf{Match.}}}} & GraphVAE & 80 & 55.7 & 66.0 & 61.6 & 26.1 \\
 & GraphVAE/imp & 40 & 56.2 & 42.0 & 75.8 & 17.9 \\
 & AAE & 40 & 13.8 & 87.1 & 66.6 & 8.0 \\
 & Ours ($\kappa = 1$) & 20 & 18.0 & 91.7 & 78.3 & 12.9 \\
 & Ours ($\kappa = -1$) & 5 & 19.1 & 50.7 & 76.5 & 7.4 \\ \bottomrule
\end{tabular}
\label{tab:mol_results}
\end{table}

\begin{figure}
    \centering
    \includegraphics[width=\linewidth, keepaspectratio=true]{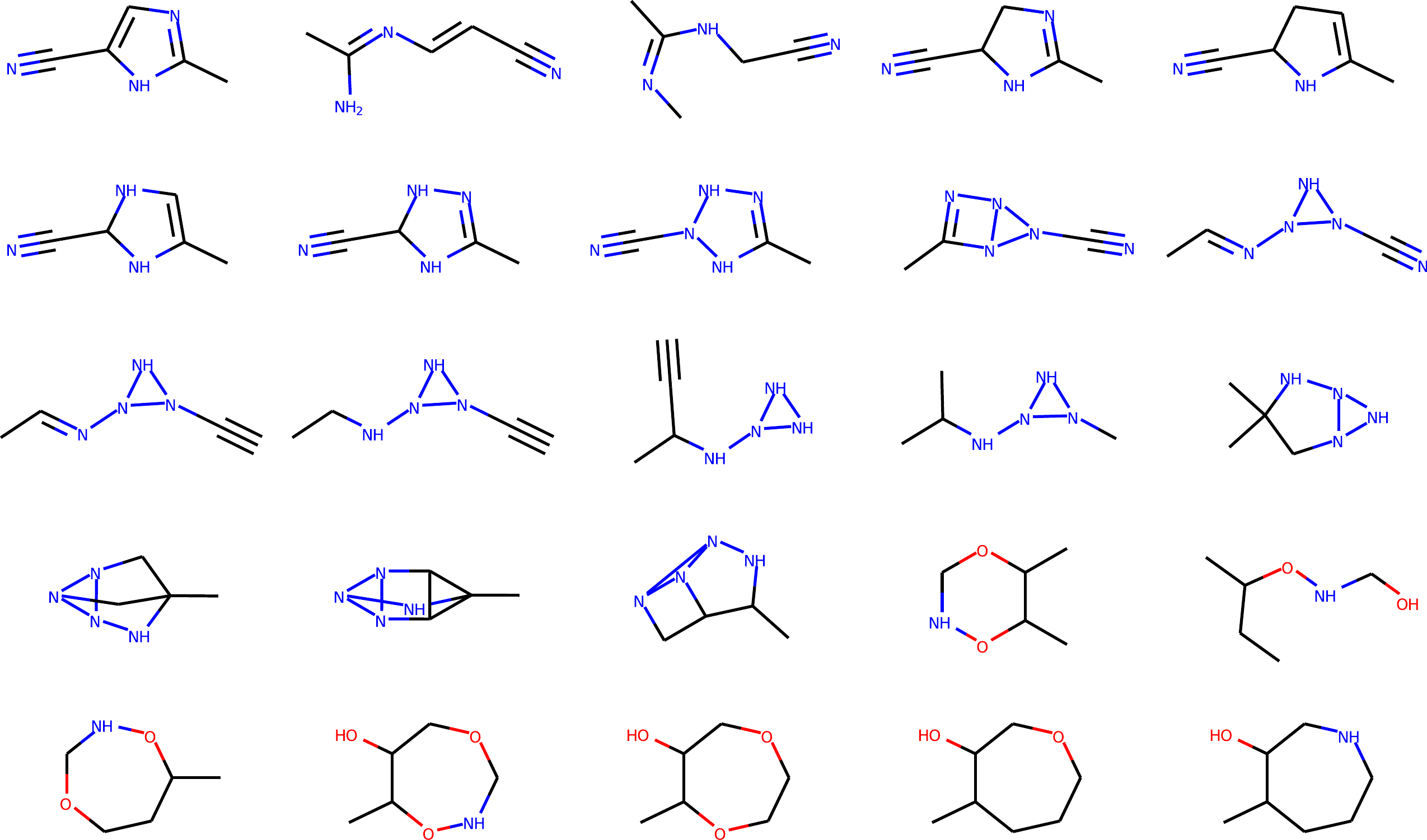}
    \caption{Traversing the latent space of a spherical CCM-AAE ($d=80$) along an equator of the sphere (samples are arranged left-to-right, top-to-bottom). We show only valid, unique, and novel molecules.}
    \label{fig:ls_mol}
\end{figure}

\paragraph{Results} A comparison of the tested models is reported in Table \ref{tab:mol_results}. We note that the spherical CCM-AAE without graph matching denotes the best performance for the \textit{joint} metric, although the performance on individual metrics is never better than the other models taken into account. All of the considered models have unbalanced performance across the three metrics (with GraphVAE being the most balanced in this regard), and we note that the CCM-AAE suffers from a low validity score. This confirms the effects of graph matching observed for GraphVAE, and explains the poor performance of the CCM-AAE when graph matching is considered (validity is halved in the spherical case). We note, however, that our model does not suffer from mode collapse, a problem commonly observed in generative adversarial networks as confirmed by the low uniqueness score of MolGAN.
Finally, sampling from the latent manifold learned with the CCM-AAE produces a smooth transition in molecule space (Figure \ref{fig:ls_mol}); however, the properties of the generated molecules are less interpretable than in the MNIST case, and we leave more focused analyses as future research.

%%%%%%%%%%%%%%%%%%%%%%%%%%%%%%%%%%%%%%%%%%%%%%%%%%%%%%%%%%%%%%%%%%%%%%%%%%%%%%
\section{Conclusions}
\label{sec:conclusions}
In this paper, we introduced an adversarial autoencoder to represent data distributions on a CCM, thus going beyond the conventional Euclidean geometry usually adopted for the latent space. The proposed method consists of jointly optimising the latent representation of CCM-AAEs to (i) match a prior with support on a CCM (i.e., a hyperspherical or hyperbolic space) and (ii) maximise the membership degree of the computed embedding vectors to the target CCM.
Experimental results on different tasks and data, ranging from images to molecules, confirm that learning representations on non-Euclidean spaces can be beneficial w.r.t.\ standard Euclidean spaces, and that the CCM-AAE shows a notable performance improvement w.r.t.\ other equivalent models in the literature.
However, the CCM-AAE shows some shortcomings w.r.t.\ simpler models (e.g., graph matching causes a severe performance drop in molecule generation), and future work might wish to address these issues by refining the architecture. Moreover, the performance of the CCM-AAE depends on the particular geometry chosen for the latent representation: it is difficult to decide \textit{a priori} which is the best choice for a given problem. To this end, it could be useful to perform an analysis of the distortion introduced by embedding the data on CCMs w.r.t. some distance function in the data space and then choosing the curvature $\kappa$ that minimises the distortion \cite{zambon2018anomaly}.

An interesting direction for future research would be to further generalise the CCM-AAE, making the curvature of the latent space a learnable parameter of the network. Moreover, we note that the CCM-AAE can be easily extended to support any arbitrary non-Euclidean manifold, via a custom prior and membership function, but we leave this possibility to future work.

\paragraph{Acknowledgements} This research is funded by the Swiss National Science Foundation project ALPSFORT (200021/172671). We gratefully acknowledge the support of NVIDIA Corporation with the donation of the Titan Xp GPU used for this research.
LL gratefully acknowledges partial support of the Canada Research Chairs program.

%%%%%%%%%%%%%%%%%%%%%%%%%%%%%%%%%%%%%%%%%%%%%%%%%%%%%%%%%%%%%%%%%%%%%%%%%%%%%%

\clearpage
\bibliography{main}

\begin{thebibliography}{29}
\expandafter\ifx\csname natexlab\endcsname\relax\def\natexlab#1{#1}\fi
\providecommand{\url}[1]{\texttt{#1}}
\providecommand{\href}[2]{#2}
\providecommand{\path}[1]{#1}
\providecommand{\DOIprefix}{doi:}
\providecommand{\ArXivprefix}{arXiv:}
\providecommand{\URLprefix}{URL: }
\providecommand{\Pubmedprefix}{pmid:}
\providecommand{\doi}[1]{\href{http://dx.doi.org/#1}{\path{#1}}}
\providecommand{\Pubmed}[1]{\href{pmid:#1}{\path{#1}}}
\providecommand{\bibinfo}[2]{#2}
\ifx\xfnm\relax \def\xfnm[#1]{\unskip,\space#1}\fi
%Type = Article
\bibitem[{Davidson et~al.(2018)Davidson, Falorsi, De~Cao, Kipf, and
  Tomczak}]{davidson2018hyperspherical}
\bibinfo{author}{T.~R. Davidson}, \bibinfo{author}{L.~Falorsi},
  \bibinfo{author}{N.~De~Cao}, \bibinfo{author}{T.~Kipf},
  \bibinfo{author}{J.~M. Tomczak},
\newblock \bibinfo{title}{Hyperspherical variational auto-encoders},
\newblock \bibinfo{journal}{34th Conference on Uncertainty in Artificial
  Intelligence (UAI-18)}  (\bibinfo{year}{2018}).
%Type = Inproceedings
\bibitem[{Nickel and Kiela(2017)}]{nickel2017poincare}
\bibinfo{author}{M.~Nickel}, \bibinfo{author}{D.~Kiela},
\newblock \bibinfo{title}{Poincar{\'e} embeddings for learning hierarchical
  representations},
\newblock in: \bibinfo{booktitle}{Advances in Neural Information Processing
  Systems}, \bibinfo{year}{2017}, pp. \bibinfo{pages}{6341--6350}.
%Type = Inproceedings
\bibitem[{Ganea et~al.(2018)Ganea, B{\'e}cigneul, and
  Hofmann}]{ganea2018hyperbolic}
\bibinfo{author}{O.~Ganea}, \bibinfo{author}{G.~B{\'e}cigneul},
  \bibinfo{author}{T.~Hofmann},
\newblock \bibinfo{title}{Hyperbolic neural networks},
\newblock in: \bibinfo{booktitle}{Advances in Neural Information Processing
  Systems}, \bibinfo{year}{2018}, pp. \bibinfo{pages}{5350--5360}.
%Type = Article
\bibitem[{Gulcehre et~al.(2018)Gulcehre, Denil, Malinowski, Razavi, Pascanu,
  Hermann, Battaglia, Bapst, Raposo, Santoro et~al.}]{gulcehre2018hyperbolic}
\bibinfo{author}{C.~Gulcehre}, \bibinfo{author}{M.~Denil},
  \bibinfo{author}{M.~Malinowski}, \bibinfo{author}{A.~Razavi},
  \bibinfo{author}{R.~Pascanu}, \bibinfo{author}{K.~M. Hermann},
  \bibinfo{author}{P.~Battaglia}, \bibinfo{author}{V.~Bapst},
  \bibinfo{author}{D.~Raposo}, \bibinfo{author}{A.~Santoro}, et~al.,
\newblock \bibinfo{title}{Hyperbolic attention networks},
\newblock \bibinfo{journal}{arXiv preprint arXiv:1805.09786}
  (\bibinfo{year}{2018}).
%Type = Article
\bibitem[{Grattarola et~al.(2018)Grattarola, Zambon, Alippi, and
  Livi}]{grattarola2018learning}
\bibinfo{author}{D.~Grattarola}, \bibinfo{author}{D.~Zambon},
  \bibinfo{author}{C.~Alippi}, \bibinfo{author}{L.~Livi},
\newblock \bibinfo{title}{Change detection in graph streams by learning graph
  embeddings on constant-curvature manifolds},
\newblock \bibinfo{journal}{arXiv preprint arXiv:1805.06299}
  (\bibinfo{year}{2018}).
%Type = Article
\bibitem[{Wilson et~al.(2014)Wilson, Hancock, P\c{e}kalska, and
  Duin}]{wilson2014spherical}
\bibinfo{author}{R.~C. Wilson}, \bibinfo{author}{E.~R. Hancock},
  \bibinfo{author}{E.~P\c{e}kalska}, \bibinfo{author}{R.~P.~W. Duin},
\newblock \bibinfo{title}{Spherical and hyperbolic {E}mbeddings of data},
\newblock \bibinfo{journal}{IEEE Transactions on Pattern Analysis and Machine
  Intelligence} \bibinfo{volume}{36} (\bibinfo{year}{2014})
  \bibinfo{pages}{2255--2269}. \DOIprefix\doi{10.1109/TPAMI.2014.2316836}.
%Type = Article
\bibitem[{Bronstein et~al.(2017)Bronstein, Bruna, LeCun, Szlam, and
  Vandergheynst}]{bronstein2016geometric}
\bibinfo{author}{M.~M. Bronstein}, \bibinfo{author}{J.~Bruna},
  \bibinfo{author}{Y.~LeCun}, \bibinfo{author}{A.~Szlam},
  \bibinfo{author}{P.~Vandergheynst},
\newblock \bibinfo{title}{Geometric deep learning: {G}oing beyond {E}uclidean
  data},
\newblock \bibinfo{journal}{IEEE Signal Processing Magazine}
  \bibinfo{volume}{34} (\bibinfo{year}{2017}) \bibinfo{pages}{18--42}.
  \DOIprefix\doi{10.1109/MSP.2017.2693418}.
%Type = Article
\bibitem[{Korman(2018)}]{korman2018autoencoding}
\bibinfo{author}{E.~O. Korman},
\newblock \bibinfo{title}{Autoencoding topology},
\newblock \bibinfo{journal}{arXiv preprint arXiv:1803.00156}
  (\bibinfo{year}{2018}).
%Type = Inproceedings
\bibitem[{Shao et~al.(2018)Shao, Kumar, and
  Thomas~Fletcher}]{shao2017riemannian}
\bibinfo{author}{H.~Shao}, \bibinfo{author}{A.~Kumar},
  \bibinfo{author}{P.~Thomas~Fletcher},
\newblock \bibinfo{title}{The riemannian geometry of deep generative models},
\newblock in: \bibinfo{booktitle}{Proceedings of the IEEE Conference on
  Computer Vision and Pattern Recognition Workshops}, \bibinfo{year}{2018}, pp.
  \bibinfo{pages}{315--323}.
%Type = Article
\bibitem[{Arvanitidis et~al.(2017)Arvanitidis, Hansen, and
  Hauberg}]{arvanitidis2017latent}
\bibinfo{author}{G.~Arvanitidis}, \bibinfo{author}{L.~K. Hansen},
  \bibinfo{author}{S.~Hauberg},
\newblock \bibinfo{title}{Latent space oddity: on the curvature of deep
  generative models},
\newblock \bibinfo{journal}{arXiv preprint arXiv:1710.11379}
  (\bibinfo{year}{2017}).
%Type = Inproceedings
\bibitem[{Xu and Durrett(2018)}]{2018arXiv180810805X}
\bibinfo{author}{J.~Xu}, \bibinfo{author}{G.~Durrett},
\newblock \bibinfo{title}{Spherical latent spaces for stable variational
  autoencoders},
\newblock in: \bibinfo{booktitle}{Proceedings of the 2018 Conference on
  Empirical Methods in Natural Language Processing}, \bibinfo{year}{2018}.
%Type = Inproceedings
\bibitem[{Zambon et~al.(2018)Zambon, Livi, and Alippi}]{zambon2018anomaly}
\bibinfo{author}{D.~Zambon}, \bibinfo{author}{L.~Livi},
  \bibinfo{author}{C.~Alippi},
\newblock \bibinfo{title}{Anomaly and change detection in graph streams through
  constant-curvature manifold embeddings},
\newblock in: \bibinfo{booktitle}{International Joint Conference on Neural
  Networks}, \bibinfo{address}{Rio de Janeiro, Brasil}, \bibinfo{year}{2018},
  pp. \bibinfo{pages}{1--8}.
%Type = Book
\bibitem[{Fisher et~al.(1993)Fisher, Lewis, and
  Embleton}]{fisher1993statistical}
\bibinfo{author}{N.~I. Fisher}, \bibinfo{author}{T.~Lewis},
  \bibinfo{author}{B.~J. Embleton}, \bibinfo{title}{Statistical analysis of
  spherical data}, \bibinfo{publisher}{Cambridge university press},
  \bibinfo{year}{1993}.
%Type = Inproceedings
\bibitem[{Makhzani et~al.(2016)Makhzani, Shlens, Jaitly, and
  Goodfellow}]{makhzani2016adversarial}
\bibinfo{author}{A.~Makhzani}, \bibinfo{author}{J.~Shlens},
  \bibinfo{author}{N.~Jaitly}, \bibinfo{author}{I.~Goodfellow},
\newblock \bibinfo{title}{Adversarial autoencoders},
\newblock in: \bibinfo{booktitle}{International Conference on Learning
  Representations}, \bibinfo{year}{2016}.
%Type = Inproceedings
\bibitem[{Goodfellow et~al.(2014)Goodfellow, Pouget-Abadie, Mirza, Xu,
  Warde-Farley, Ozair, Courville, and Bengio}]{goodfellow2014generative}
\bibinfo{author}{I.~Goodfellow}, \bibinfo{author}{J.~Pouget-Abadie},
  \bibinfo{author}{M.~Mirza}, \bibinfo{author}{B.~Xu},
  \bibinfo{author}{D.~Warde-Farley}, \bibinfo{author}{S.~Ozair},
  \bibinfo{author}{A.~Courville}, \bibinfo{author}{Y.~Bengio},
\newblock \bibinfo{title}{Generative adversarial nets},
\newblock in: \bibinfo{booktitle}{Advances in Neural Information Processing
  Systems}, \bibinfo{year}{2014}, pp. \bibinfo{pages}{2672--2680}.
%Type = Inproceedings
\bibitem[{Straub et~al.(2015)Straub, Chang, Freifeld, and {Fisher
  III}}]{straub2015dirichlet}
\bibinfo{author}{J.~Straub}, \bibinfo{author}{J.~Chang},
  \bibinfo{author}{O.~Freifeld}, \bibinfo{author}{J.~{Fisher III}},
\newblock \bibinfo{title}{A {D}irichlet process mixture model for spherical
  data},
\newblock in: \bibinfo{booktitle}{Proceedings of the 18th International
  Conference on Artificial Intelligence and Statistics},
  volume~\bibinfo{volume}{38}, \bibinfo{address}{San Diego, CA, USA},
  \bibinfo{year}{2015}, pp. \bibinfo{pages}{930--938}.
%Type = Inproceedings
\bibitem[{Simonovsky and Komodakis(2018)}]{simonovsky2018graphvae}
\bibinfo{author}{M.~Simonovsky}, \bibinfo{author}{N.~Komodakis},
\newblock \bibinfo{title}{Graph{VAE}: Towards generation of small graphs using
  variational autoencoders},
\newblock in: \bibinfo{booktitle}{27th International Conference on Artificial
  Neural Networks}, \bibinfo{year}{2018}, pp. \bibinfo{pages}{412--422}.
  \DOIprefix\doi{10.1007/978-3-030-01418-6\_41}.
%Type = Article
\bibitem[{De~Cao and Kipf(2018)}]{de2018molgan}
\bibinfo{author}{N.~De~Cao}, \bibinfo{author}{T.~Kipf},
\newblock \bibinfo{title}{Molgan: An implicit generative model for small
  molecular graphs},
\newblock \bibinfo{journal}{arXiv preprint arXiv:1805.11973}
  (\bibinfo{year}{2018}).
%Type = Inproceedings
\bibitem[{Kingma et~al.(2014)Kingma, Mohamed, Rezende, and
  Welling}]{kingma2014semi}
\bibinfo{author}{D.~P. Kingma}, \bibinfo{author}{S.~Mohamed},
  \bibinfo{author}{D.~J. Rezende}, \bibinfo{author}{M.~Welling},
\newblock \bibinfo{title}{Semi-supervised learning with deep generative
  models},
\newblock in: \bibinfo{booktitle}{Advances in Neural Information Processing
  Systems}, \bibinfo{year}{2014}, pp. \bibinfo{pages}{3581--3589}.
%Type = Inproceedings
\bibitem[{Salakhutdinov and Murray(2008)}]{salakhutdinov2008quantitative}
\bibinfo{author}{R.~Salakhutdinov}, \bibinfo{author}{I.~Murray},
\newblock \bibinfo{title}{On the quantitative analysis of deep belief
  networks},
\newblock in: \bibinfo{booktitle}{Proceedings of the 25th international
  conference on Machine learning}, \bibinfo{organization}{ACM},
  \bibinfo{year}{2008}, pp. \bibinfo{pages}{872--879}.
%Type = Article
\bibitem[{Kingma and Welling(2013)}]{kingma2013auto}
\bibinfo{author}{D.~P. Kingma}, \bibinfo{author}{M.~Welling},
\newblock \bibinfo{title}{Auto-encoding variational bayes},
\newblock \bibinfo{journal}{arXiv preprint arXiv:1312.6114}
  (\bibinfo{year}{2013}).
%Type = Article
\bibitem[{Kipf and Welling(2016)}]{kipf2016variational}
\bibinfo{author}{T.~N. Kipf}, \bibinfo{author}{M.~Welling},
\newblock \bibinfo{title}{Variational graph auto-encoders},
\newblock \bibinfo{journal}{NIPS Workshop on Bayesian Deep Learning}
  (\bibinfo{year}{2016}).
%Type = Inproceedings
\bibitem[{Kipf and Welling(2017)}]{kipf2017semi}
\bibinfo{author}{T.~N. Kipf}, \bibinfo{author}{M.~Welling},
\newblock \bibinfo{title}{Semi-supervised classification with graph
  convolutional networks},
\newblock in: \bibinfo{booktitle}{International Conference on Learning
  Representations}, \bibinfo{year}{2017}.
%Type = Inproceedings
\bibitem[{Samanta et~al.(2018)Samanta, De, Jana, Chattaraj, Ganguly, and
  Gomez-Rodriguez}]{samanta2018NeVAEAD}
\bibinfo{author}{B.~Samanta}, \bibinfo{author}{A.~De},
  \bibinfo{author}{G.~Jana}, \bibinfo{author}{P.~K. Chattaraj},
  \bibinfo{author}{N.~Ganguly}, \bibinfo{author}{M.~Gomez-Rodriguez},
\newblock \bibinfo{title}{Nevae: A deep generative model for molecular graphs},
\newblock \bibinfo{year}{2018}.
%Type = Inproceedings
\bibitem[{Simonovsky and Komodakis(2017)}]{simonovsky2017ecc}
\bibinfo{author}{M.~Simonovsky}, \bibinfo{author}{N.~Komodakis},
\newblock \bibinfo{title}{Dynamic edge-conditioned filters in convolutional
  neural networks on graphs},
\newblock in: \bibinfo{booktitle}{Conference on Computer Vision and Pattern
  Recognition}, \bibinfo{year}{2017}.
%Type = Inproceedings
\bibitem[{Li et~al.(2016)Li, Tarlow, Brockschmidt, and Zemel}]{li2015gated}
\bibinfo{author}{Y.~Li}, \bibinfo{author}{D.~Tarlow},
  \bibinfo{author}{M.~Brockschmidt}, \bibinfo{author}{R.~Zemel},
\newblock \bibinfo{title}{Gated graph sequence neural networks},
\newblock in: \bibinfo{booktitle}{International Conference on Learning
  Representations}, \bibinfo{year}{2016}.
%Type = Inproceedings
\bibitem[{Cho et~al.(2014)Cho, Sun, Duchenne, and Ponce}]{cho2014finding}
\bibinfo{author}{M.~Cho}, \bibinfo{author}{J.~Sun},
  \bibinfo{author}{O.~Duchenne}, \bibinfo{author}{J.~Ponce},
\newblock \bibinfo{title}{Finding matches in a haystack: A max-pooling strategy
  for graph matching in the presence of outliers},
\newblock in: \bibinfo{booktitle}{Proceedings of the IEEE Conference on
  Computer Vision and Pattern Recognition}, \bibinfo{year}{2014}, pp.
  \bibinfo{pages}{2083--2090}.
%Type = Article
\bibitem[{G{\'o}mez-Bombarelli et~al.(2018)G{\'o}mez-Bombarelli, Wei, Duvenaud,
  Hern{\'a}ndez-Lobato, S{\'a}nchez-Lengeling, Sheberla, Aguilera-Iparraguirre,
  Hirzel, Adams, and Aspuru-Guzik}]{gomez2018automatic}
\bibinfo{author}{R.~G{\'o}mez-Bombarelli}, \bibinfo{author}{J.~N. Wei},
  \bibinfo{author}{D.~Duvenaud}, \bibinfo{author}{J.~M. Hern{\'a}ndez-Lobato},
  \bibinfo{author}{B.~S{\'a}nchez-Lengeling}, \bibinfo{author}{D.~Sheberla},
  \bibinfo{author}{J.~Aguilera-Iparraguirre}, \bibinfo{author}{T.~D. Hirzel},
  \bibinfo{author}{R.~P. Adams}, \bibinfo{author}{A.~Aspuru-Guzik},
\newblock \bibinfo{title}{Automatic chemical design using a data-driven
  continuous representation of molecules},
\newblock \bibinfo{journal}{ACS central science} \bibinfo{volume}{4}
  (\bibinfo{year}{2018}) \bibinfo{pages}{268--276}.
%Type = Inproceedings
\bibitem[{Kusner et~al.(2017)Kusner, Paige, and
  Hern{\'a}ndez-Lobato}]{kusner2017grammar}
\bibinfo{author}{M.~J. Kusner}, \bibinfo{author}{B.~Paige},
  \bibinfo{author}{J.~M. Hern{\'a}ndez-Lobato},
\newblock \bibinfo{title}{Grammar variational autoencoder},
\newblock in: \bibinfo{booktitle}{International Conference on Machine
  Learning}, \bibinfo{year}{2017}, pp. \bibinfo{pages}{1945--1954}.

\end{thebibliography}
\bibliographystyle{elsarticle-num-names}
\end{document}